\documentclass[journal,comsoc]{IEEEtran}
\usepackage[T1]{fontenc}
\usepackage{cite}

\ifCLASSINFOpdf
   \usepackage[pdftex]{graphicx}
\else
   \usepackage[dvips]{graphicx}
\fi

\usepackage{amsmath}
\usepackage[cmintegrals]{newtxmath}

\usepackage[numbers]{natbib}
\usepackage{booktabs}
\usepackage{multirow}
\usepackage{multicol}
\usepackage{amsmath}
\usepackage{algpseudocode}
\usepackage{mathtools}
\usepackage[linesnumbered,ruled,vlined]{algorithm2e}

\mathchardef\mhyphen="2D

\begin{document}
%
\title{Human-Inspired Continuous Learning of Internal Reasoning Processes: Learning How to Think for Adaptive AI Systems}

\author{Hong~Su
\IEEEcompsocitemizethanks{\IEEEcompsocthanksitem H. Su is with the School of Computer Science, Chengdu University of information Technology, Chengdu, China.\\
 E-mail: suguest@126.com. \\
\protect\\
}
\thanks{}}

\markboth{Journal of \LaTeX\ Class Files,~Vol.~14, No.~8, August~2015}%
{Shell \MakeLowercase{\textit{et al.}}: Bare Demo of IEEEtran.cls for IEEE Communications Society Journals}
%

\maketitle

\begin{abstract}

Learning internal reasoning processes is crucial for developing AI systems capable of sustained adaptation in dynamic real-world environments. However, most existing approaches primarily emphasize learning task-specific outputs or static knowledge representations, while overlooking the continuous refinement of internal reasoning structures, action scheduling policies, and learning mechanisms themselves.

In this paper, we propose a human-inspired continuous learning framework that unifies reasoning, action, reflection, and verification within a sequential reasoning model enhanced by parallel learning. The framework explicitly treats internal thinking processes as primary learning objects. It systematically records internal reasoning trajectories and environmental interactions as structured learning material, enabling the system to optimize not only task-level content but also the organization, scheduling, and evolution of reasoning activities. This design realizes learning alongside processing, allowing cognitive structures to improve during execution. Furthermore, the framework supports controlled replacement of predefined logic with learned procedures and introduces a hierarchical learning-to-learn mechanism that jointly adapts task-level parameters and learning strategies. As a result, the system progressively evolves its internal cognitive architecture while preserving operational stability.

Experimental results on a temperature sensor abnormality detection task show that incorporating internal-process learning reduces average runtime by 23.9\%.

\end{abstract}


\IEEEpeerreviewmaketitle

\section{Introduction}

A person does not merely accumulate knowledge but continuously refines how to think, act, and learn. When encountering new situations, humans adapt not only by recalling past information, but by reorganizing internal reasoning activities, adjusting decision schedules, and even improving their own learning strategies. This ability to modify internal cognitive processes is a fundamental source of long-term adaptability in dynamic environments.

For humans, learning \emph{how to think} is often more valuable than simply knowing isolated facts. Intelligence arises not only from stored knowledge, but from the structure and evolution of internal reasoning processes, including planning, reflection, action selection, and learning itself. Therefore, if AI systems are to achieve similar adaptability, they must be trained not only on external language materials but also on the mechanisms that govern internal cognitive activities.

In contrast, most current AI systems—particularly those based on large language models (LLMs) \cite{naveed2025comprehensive} — primarily focus on optimizing task outputs or expanding static knowledge representations. Although recent advances in agent frameworks have enabled reasoning--action loops and reflection mechanisms, the internal structure of reasoning processes is typically predefined and remains largely fixed during deployment \cite{su2026hsc}. Learning is commonly applied to model parameters or task-level outputs, while the organization, scheduling, and evolution of internal reasoning activities are rarely treated as adaptive objects.

To address this limitation, we propose a human-inspired continuous learning framework in which internal reasoning activities, action scheduling, and learning mechanisms are explicitly modeled as evolvable components. The framework integrates sequential reasoning with parallel learning, enabling the system to record internal reasoning trajectories and environmental interactions as structured learning material. Rather than treating learning as an offline phase, our approach embeds learning within the reasoning--action loop, allowing both task behavior and internal cognitive structures to improve progressively over time.

The main contributions of this paper are summarized as follows:

\begin{itemize}
    \item We propose a continuous learning framework that treats internal reasoning activities and their scheduling mechanisms as learnable objects, enabling structural adaptation of cognitive processes rather than only content optimization.

    \item We introduce a mechanism for replacing predefined execution logic with learned procedures, allowing fixed code modules to be removed once more effective learned alternatives are verified.

    \item We extend the framework to support learning-to-learn, in which the learning mechanism itself can be updated or replaced by improved strategies while maintaining operational stability.
\end{itemize}

The remainder of this paper is organized as follows. Section~II presents the proposed continuous learning model, including sequential reasoning with parallel learning, internal process modeling, and meta-level adaptation. Section~III describes the experimental verification and comparative evaluation. Finally, Section~IV concludes the paper and discusses future research directions.

\section{Related Work} \label{sec_related_work}
We review related studies from three perspectives that are most relevant to our objective: (i) LLM-based agent frameworks that interleave reasoning and action, (ii) continual learning for large language models, and (iii) supervision and evaluation of step-by-step reasoning traces.

\subsection{LLM Agents with Reasoning--Action Loops and Reflection}

Recent work has demonstrated that LLMs can be extended from pure text generators into interactive agents by explicitly interleaving reasoning with tool use and environmental actions \cite{muennighoff2025s1} \cite{su2026env}. ReAct formalizes the synergy between reasoning traces and action execution, enabling agents to query external sources and reduce hallucination in interactive settings \cite{yao2023react} . Toolformer further trains models to decide when and how to invoke external tools, shifting tool usage from purely prompt-based design toward learned tool invocation \cite{schick2023toolformer}. Reflection-based approaches enhance agent behavior by storing linguistic feedback as episodic memory and leveraging it to guide subsequent trials, as exemplified by Reflexion \cite{shinn2023reflexion} and iterative self-feedback methods such as Self-Refine \cite{madaan2023selfrefine} \cite{zhao2026retrieval}. Comprehensive surveys have also summarized the rapidly expanding landscape of LLM-based autonomous agents and tool-augmented systems \cite{shi2025continual}.

While these approaches improve task performance through better prompting strategies, memory utilization, and tool selection, they generally treat reasoning modules and interaction loops as fixed structural templates. Internal processes are often limited to predefined components such as reflection or memory retrieval, and their organization is not itself a primary learning target. In contrast, our framework treats the internal process structure—including activity scheduling, insertion and removal of reasoning modules, and replacement of predefined execution logic—as an adaptive and continuously learnable object. Moreover, internal processes are not restricted to reflection mechanisms; rather, they encompass \emph{diverse and dynamically} evolving forms of thinking that can vary across tasks and contexts. This broader view of internal reasoning as a structural and evolvable entity distinguishes our work from existing LLM agent frameworks.

\subsection{Continual Learning for Large Language Models}

Continual learning aims to enable models to adapt to evolving data distributions and new tasks without catastrophic forgetting. In the context of large language models (LLMs), continual learning has been explored across multiple stages, including continual pretraining, instruction tuning, and alignment. Existing approaches primarily focus on parameter-level updates, data replay mechanisms, regularization strategies, and modular architectures to preserve previously acquired knowledge while incorporating new information \cite{shi2025continual}.

Several works investigate efficient adaptation techniques such as parameter-efficient fine-tuning (PEFT) \cite{xu2026parameter}, adapter-based learning, and low-rank updates to incrementally extend model capabilities while controlling computational cost \cite{eschbach2024exploring}. Other studies explore memory-based methods and rehearsal strategies to mitigate forgetting during sequential task training. These approaches aim to maintain performance stability under distribution shifts and long-term deployment scenarios.

While continual learning for LLMs has made substantial progress at the parameter and data levels, most existing methods treat the internal reasoning structure of the model as fixed. Adaptation is typically achieved by updating model weights or introducing auxiliary modules, rather than modifying the organization, scheduling, or structural composition of reasoning activities during inference. In contrast, our framework emphasizes continual adaptation of internal reasoning processes themselves, including dynamic restructuring of reasoning steps and replacement of predefined execution logic. Thus, our work complements parameter-centric continual learning by introducing structural-level adaptation within the reasoning–action loop.

\subsection{Process supervision and reasoning-trace evaluation.}
Another related direction is improving and assessing step-by-step reasoning, where models are trained or evaluated on intermediate reasoning steps rather than only final answers. Process supervision has been shown to improve mathematical reasoning by rewarding correct intermediate steps and releasing step-level supervision datasets \cite{luo2024improve}. More recently, surveys summarize evaluation practices and challenges for reasoning traces \cite{lee2025reasontracesurvey}, and reflection-reinforced self-training explores using reflective feedback signals to improve agentic behavior \cite{dou2024reflectionselftraining}.

These works mainly target correctness and interpretability of reasoning traces for specific tasks (e.g., math or QA), while our setting emphasizes continuous learning of internal process organization and scheduling in environment-interactive tasks, where correctness is grounded by action outcomes and verification feedback rather than only textual supervision.

\section{General Continuous Learning with Internal Human-Simulation Processes}

Human intelligence arises not only from the accumulation of knowledge, but more fundamentally from the diversity of thinking methods used to acquire, apply, and refine that knowledge. In many cases, learning methods are more important than the knowledge itself. However, most existing learning paradigms primarily focus on external materials or final outputs, while the internal thinking processes and action-generation mechanisms are rarely learned explicitly or jointly.

Learning internal processes is crucial because it enables an AI system to train and refine its own reasoning mechanisms, rather than merely memorizing the products of reasoning (e.g., language outputs). This capability is particularly important in real-world environments, where an AI system must handle diverse and previously unseen situations. For example, when lifting a kitchen knife for the first time, the required force cannot be fully specified in advance. Through interaction, trial, and feedback, a human gradually learns the appropriate strength. By modeling and learning such internal simulation and action-adjustment processes, an AI system can similarly acquire practical skills through experience rather than relying solely on predefined rules or static data.

\subsection{Learning Along with Processing: On-Time Acquisition of Reasoning and Actions}

In this paper, we advocate a learning paradigm in which all observable and internal processes are learned \emph{on time}, including thinking activities, their semantic content, and interactions with the external environment. Instead of postponing learning to offline stages or relying on external supervision, the proposed system learns within a relatively short time window through its own processing experience.

\begin{figure}[htpb]
    \centering
    \includegraphics[width = 3.5in]{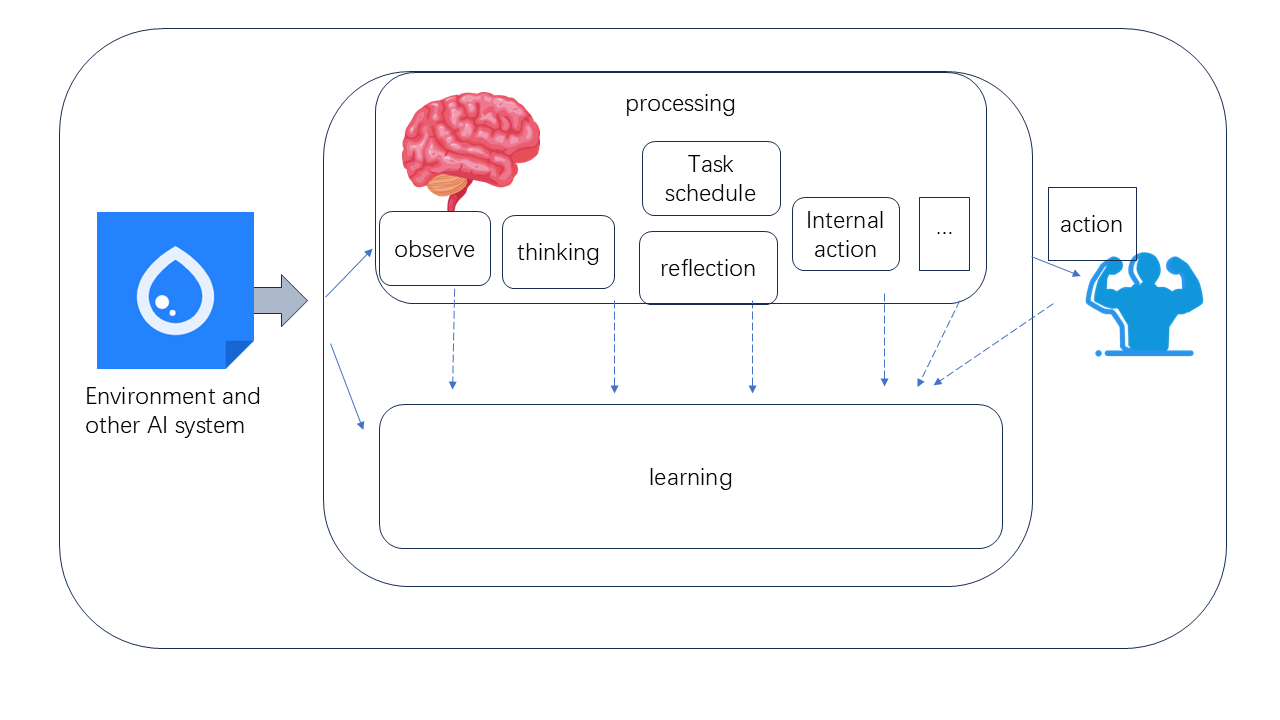}
    \caption{Learning integrated with the full human-like processing pipeline}
    \label{fig_architecture}
\end{figure}

The core idea is that every processing step executed by the AI system is recorded and immediately provided to the learning module. This mechanism is inspired by human cognition: when a human performs a task, repeated exposure to the same process leads to increased familiarity and improved performance. As illustrated in Fig.~\ref{fig_architecture}, learning is tightly coupled with processing itself, forming a unified loop of execution and adaptation.

The proposed framework requires an AI system (which can be implemented using a large language model) equipped with a human-simulation computation mechanism~\cite{su2026env}. Such a system not only performs reasoning from input prompts to output responses, but also incorporates reflection, action execution, and learning as first-class components. Specifically, the system consists of an LLM core, external memory, an action executor, and auxiliary modules that support continuous learning and intermediate coordination between reasoning and action execution.

In this setting, learning is not an isolated stage but is embedded within the complete reasoning--action cycle, as abstractly expressed in Eq.~\eqref{eq_learning_on_brain_activity}.

\begin{equation}
\label{eq_learning_on_brain_activity}
\text{Logic}
\xrightarrow{\text{cognitive process}}
\left\{
\begin{aligned}
&\text{Thinking} \rightarrow \text{Action} \rightarrow \cdots \\
&\hphantom{123}\downarrow  \hphantom{Thinking} \downarrow \hphantom{Action} \downarrow \\
&\text{Learning} \rightarrow  \text{Learning} \rightarrow  \text{Learning}
\end{aligned}
\right.
\end{equation}

Equation~\eqref{eq_learning_on_brain_activity} emphasizes that learning is triggered and driven by internal reasoning activities rather than being limited to external supervision signals. By directly coupling thinking and learning, the system can refine its internal processes while tasks are being executed.

\subsubsection{Sequential Reasoning with Parallel Learning}

We assume that reasoning and decision-making occur as a sequential process composed of multiple steps during interaction with the environment. This assumption follows human cognitive constraints: except for learning, which can occur asynchronously, most reasoning activities are performed in sequence. Accordingly, this paper adopts a sequence model for reasoning and action execution; parallel reasoning models are left for future work.

Let the system state at time $t$ be denoted by $s_t$, which includes internal representations and observed environmental information. The reasoning--action process can be modeled as a sequential transition:
\begin{equation}
\label{eq_seq_transition}
s_{t+1} = f(s_t, a_t),
\end{equation}
where $a_t$ is the action (which may include internal reasoning steps or external operations), and $f(\cdot)$ represents the environment-coupled state transition function.

Under this assumption, the AI system addresses internal and external challenges through a sequence of reasoning steps that collectively simulate human problem-solving behavior. Let the reasoning trajectory over a task horizon $T$ be defined as
\begin{equation}
\label{eq_reasoning_sequence}
\mathcal{R} = \{(s_0, a_0), (s_1, a_1), \ldots, (s_T, a_T)\}.
\end{equation}

Learning operates in parallel with this sequence by updating the internal model parameters $\theta$ based on the observed trajectory $\mathcal{R}$:
\begin{equation}
\label{eq_parallel_learning}
\theta \leftarrow \mathcal{L}(\theta, \mathcal{R}),
\end{equation}
where $\mathcal{L}(\cdot)$ denotes the learning operator. Importantly, the update in Eq.~\eqref{eq_parallel_learning} does not interrupt the execution of Eq.~\eqref{eq_seq_transition}, which motivates the term \emph{parallel learning model}.

In each interaction cycle, the system performs not only internal reasoning but also external actions that affect the environment. While actions and reasoning steps are executed sequentially, learning observes these processes concurrently and updates the internal model without interrupting task execution.

The internal process can be \emph{generalized} as a set of reasoning activities $\mathcal{A} = \{a^{(1)}, a^{(2)}, \ldots, a^{(K)}\}$, each associated with specific semantic content and scheduled at different time points. These activities are abstracted, recorded, and transformed into learning materials, as shown in Eq.~\eqref{eq_gen_rea_act}. Through training on accumulated materials, the system extracts reusable patterns of internal reasoning that can be applied in future tasks.

\begin{equation}
\label{eq_gen_rea_act}
\mathcal{A}
\xrightarrow{\mathcal{T}}
\mathcal{M}
\xrightarrow{\mathcal{U}}
\mathcal{G}
\end{equation}

 , where $\mathcal{A}$ denotes the set of reasoning activities,
$\mathcal{M}$ represents the recorded learning material,
$\mathcal{T}$ is the transfer operator,
$\mathcal{U}$ is the training process,
and $\mathcal{G}$ denotes the extracted generalized reasoning rules.

The generalization above provides an abstraction of the proposed model: all internal operations, including action execution and verification, are unified as state-dependent cognitive operators. This unification enables consistent recording, learning, and structural restructuring across heterogeneous activities.

Consequently, learning in this framework targets not only \emph{what} to do (execution content), but also \emph{how} and \emph{when} to do it (reasoning activities and their scheduling). Formally, let $\pi(a \mid s; \theta)$ denote the policy that schedules reasoning activities based on the current state $s$. Learning therefore updates both the execution parameters and the scheduling policy $\pi$, enabling insertion, removal, refinement, or reorganization of reasoning steps.

Importantly, \emph{the internal process is not static}. The segmentation and organization of reasoning steps are guided by ongoing cognitive evaluation and previously accumulated experience. The introduction of new reasoning steps corresponds to learning new cognitive operations, such as adding reflection after action execution or introducing additional knowledge acquisition stages. Moreover, structurally similar reasoning steps may be merged or reorganized to improve efficiency. Updates may therefore involve replacing existing steps with more effective ones, refining their execution content, or restructuring their arrangement.

From a structural perspective, these adaptations modify the reasoning trajectory $\mathcal{R}$ through state-dependent transformation operators. Such transformations preserve sequential consistency while enabling higher-level structural evolution of internal cognitive processes.

\subsubsection{Procedure}

The overall procedure records each processing step in its execution order and continuously feeds these records to the learning system. The recorded information includes sensed external inputs (analogous to human vision, hearing, and tactile perception), internal reasoning activities (e.g., prompts provided to the LLM and intermediate outputs), and executed actions. Together, these elements constitute the learning material.

A key characteristic of the proposed framework is that learning materials are generated and organized in temporal sequence, reflecting the order in which the AI system experiences the environment. By learning from temporally ordered records, the system naturally acquires procedural knowledge, which enables it to handle previously unseen tasks by reusing and adapting learned process patterns.

Internal reasoning activities and externally sensed information are processed and recorded during execution and passed to the learning module either immediately or during later reflection stages. Learning may therefore occur on time or during delayed retrospection, while still remaining parallel to task execution. We refer to this mechanism as \emph{parallel learning}, as learning does not interrupt the reasoning–action loop.

An important capability of this framework is the \emph{restoration of internal thinking activities}. Since internal processes are explicitly recorded, they can be reconstructed from logs and replayed when needed. In this work, such records are implemented as activity logs; more biologically inspired representations, such as neural pulse sequences, are also possible in principle. This restoration capability enables controlled analysis, reuse, and refinement of learned reasoning procedures.

After learning, the system resumes task execution by collecting current task states and environmental information and providing them to the LLM. The LLM may generate instructions for the ongoing task, propose new tasks at predefined intervals, or trigger urgent tasks in response to environmental changes. These outputs are subsequently executed by the action module and recorded again, forming a closed learning–execution loop.

\subsubsection{Learning from Outer Sources via Simulation}

Humans can learn not only from direct experience but also from external sources such as books, demonstrations, or instruction by others. We refer to this form of learning as learning from outer sources. To simulate this process, the proposed framework integrates external information into the internal reasoning pipeline through simulation.

The core idea is to treat external information as a form of internally simulated experience. External descriptions or demonstrations are processed by the system’s internal reasoning modules and compared with existing internal knowledge. If inconsistencies are detected, they are corrected and explicitly marked for emphasized learning. The entire reasoning process induced by the external source is then recorded in the same manner as internally generated experiences.

When external learning involves actions (e.g., manipulating an object), the system maintains a simulated action interface that models the action execution without invoking real-world effects. This allows the system to learn action structures, parameters, and expected outcomes safely. Through such simulated interactions, the AI system can acquire new skills and reasoning strategies before deploying them in real environments.

\subsection{Replacement of Fixed Code through Learning} \label{sec_replace_fix_code}

The proposed framework initially includes a set of fixed, hand-designed modules that mimic human cognitive processes, such as thinking, reflection, action execution, learning, verification, and task scheduling. These modules provide a safe and interpretable starting point for system operation. However, they are not assumed to be optimal or permanent.

As learning progresses, the system continuously evaluates whether learned procedures outperform the original hard-coded mechanisms. If a learned procedure demonstrates superior performance and stability under verification, it may replace the corresponding fixed code. In this way, the system gradually transitions from predefined logic to experience-driven behavior.

The replacement process can be summarized as:
\begin{quote}
initial fixed modules (including learning and verification) $\rightarrow$ learning of full reasoning processes $\rightarrow$ validated learned procedures replace fixed modules.
\end{quote}

Replacing fixed code introduces potential risks, particularly if a learned procedure is incorrect or compromises system safety. To mitigate these risks, several safeguards are incorporated:
\begin{itemize}
    \item \textbf{Verification before replacement:} learned procedures are thoroughly tested under simulated and controlled conditions before deployment.
    \item \textbf{Fallback and recovery mechanisms:} the system retains basic reboot and recovery functions, allowing it to revert to stable states when critical failures are detected (e.g., loss of power or sensor malfunction).
    \item \textbf{Reasoned replacement decisions:} prior to replacement, the LLM evaluates whether the learned procedure is sufficiently reliable to substitute the original module.
\end{itemize}

These constraints ensure that the gradual removal of fixed code improves adaptability while maintaining robustness and safety.

\subsection{Learning Scope: Internal and External Information Acquisition}

The learning scope of the proposed framework includes all information sensed by the AI system, together with the internal reasoning processes triggered by such information. Learning is therefore driven not only by external observations but also by the system’s internal cognitive activities.

\subsubsection{Internal Processes}

Human cognition involves multiple internal processes, including observation, reasoning, reflection, learning, action execution, and verification~\cite{su2026hsc}. In the proposed framework, all such internal processes are treated as learnable components, with the objective of automatically improving both their execution and coordination over time.

These internal processes include all activities generated by the system’s reasoning core while interacting with the environment. They are not limited to explicit reasoning steps, but also encompass reflection, verification, reasoning strategy selection, and the scheduling of these activities. For example, when a human repeatedly performs reasoning for similar problems, not only the reasoning content but also the timing and invocation of reasoning are gradually optimized. Humans often recall and reflect automatically based on prior experience; such behavior can be viewed as a learned task-scheduling mechanism.

Human problem-solving typically follows identifiable stages: recognizing a task or question, selecting a method, executing actions, and evaluating outcomes. These stages are mirrored in the proposed LLM-based AI system, where reasoning is realized through LLM prompts and their ordered execution. Taking reflection as an example, reflection is triggered by issuing specific prompts to the LLM after task execution.

Two aspects are critical in this process: the prompt content and the prompt sequence. Since internal activities are recorded in temporal order, learning naturally preserves their sequential structure. This enables the system to learn not only which prompts to use, but also when they should be issued relative to task execution.

Another important factor is scheduling. The decision of when to perform reasoning, reflection, or learning is itself learned from experience. As internal activities are recorded and optimized over time, the system can autonomously initiate appropriate internal tasks when resources are available. This behavior simulates human cognition, where reflection and reasoning are often triggered automatically based on prior learning rather than explicit external commands.

\subsubsection{Learning from External Environment and Other AI Systems}

All environmental information sensed by the AI system, including system states, environmental feedback, and outcome signals, is incorporated into the learning process. Such information serves both as training material and as feedback for evaluating the correctness of reasoning and action decisions. Through repeated interaction, the system learns to associate specific environmental conditions with appropriate internal reasoning patterns and actions.

In addition to direct environmental sensing, information obtained from other AI systems is also treated as a learning source. Observing the behavior, strategies, or outcomes of other AI systems enables knowledge transfer and accelerates the formation of effective reasoning structures. This mirrors human social learning, where individuals learn not only from personal experience but also from observing others.

The reasoning processes applied to external materials are themselves learnable. For example, when a human observes a harmful object, the object is identified as dangerous, and avoidance behavior is learned. Subsequently, when similar features or characteristics are perceived, appropriate actions are triggered automatically. In the proposed framework, such behavior is learned by associating external observations with internal reasoning and action responses.

Furthermore, new reasoning activities can be introduced to analyze external observations. For instance, when encountering unfamiliar environmental features, the system may learn to invoke additional reasoning steps, such as causal analysis or explanation generation, based on prior learning from books or other knowledge sources.

\subsubsection{Interaction with the Outer Environment}

Interaction with the external environment typically involves multiple steps and continuous-valued parameters, such as action strength, duration, or direction. Effective interaction therefore requires the system to learn not only discrete action choices but also how to adjust these parameters based on environmental feedback.

When such parameters are sensed, recorded, and learned, the system gradually acquires the ability to execute actions automatically with appropriate magnitudes and timing. Over time, this may lead to the emergence of new composite actions or additional internal reasoning processes that better adapt behavior to complex environmental dynamics.

Through continuous sensing, interaction, and learning, the system refines both its internal reasoning structure and its external action execution, enabling robust adaptation in real-world environments.

\subsection{On-Time and Continuous Learning}

Learning in the proposed framework is both \emph{on-time} and \emph{continuous}. Information generated during system operation is fed into the learning process as tasks are executed, enabling timely adaptation. Learning is not restricted to isolated phases but persists throughout the entire lifetime of the AI system.

\subsubsection{On-Time Learning}

The framework encourages learning to occur alongside internal reasoning and action execution. However, since fine-tuning or updating large language models may require non-negligible computational time, learning is not forced to occur immediately at every step. Instead, the system adopts a strategy inspired by human cognition, where learning and reflection are often deferred to appropriate periods, such as daily review or retrospective summarization.

Accordingly, when computational resources are available or the system enters idle periods, past activities are revisited, summarized, and used for model refinement. This mechanism enables timely learning without disrupting real-time task execution, while still preserving the temporal relevance of experience.

\subsubsection{Continuous Learning}

Learning is also continuous across extended periods of operation. As time progresses, the system repeatedly revisits previously encountered materials to reinforce familiar patterns, while allocating greater attention to newly observed behaviors or reasoning processes.

In particular, unfamiliar methods or behaviors with high uncertainty ~\cite{su2026max} or error are prioritized for learning. When such cases are detected, the system actively initiates additional exploration or interaction to reduce uncertainty. Through this process, the system incrementally refines its reasoning and action strategies, ensuring long-term adaptability.

\subsection{Learning to Learn}

Beyond acquiring task-specific knowledge, the proposed framework supports \emph{learning to learn}, in which the learning mechanisms themselves are treated as adaptive components. Rather than being fixed, the processes that govern how information is collected, organized, and used for model updates can be modified over time.

This capability is closely related to the replacement of fixed code discussed in Section~\ref{sec_replace_fix_code}. When newly learned reasoning or learning procedures demonstrate superior performance and pass verification, they may replace existing predefined logic. Consequently, not only reasoning steps but also the learning workflow itself becomes subject to structural evolution.

The scope of adaptation includes learning inputs (e.g., internal records and externally sensed information), prompt construction strategies for the LLM, fine-tuning procedures, and even the selection of the underlying language model. For instance, if a new LLM is adopted, knowledge encoded in the previous model can be preserved through transfer, migration, or distillation mechanisms to ensure continuity.

This design mirrors human cognitive development, where foundational knowledge supports higher-level abstraction and meta-learning. Previously learned representations serve as prerequisites for improving learning efficiency and generalization ability. Over time, the system refines not only its task-solving strategies but also its strategies for acquiring and consolidating knowledge.

To maintain robustness, legacy learning mechanisms are temporarily retained when new learning strategies are introduced. If performance degradation or instability is detected, the system can revert to previously validated mechanisms. This layered adaptation strategy enables progressive enhancement of learning capability while preserving operational safety and stability.

\subsection{Application and Verification of Thinking}

In real-world environments, actions and reasoning processes often involve multiple steps and uncertain outcomes. Simply generating an internal conclusion is insufficient; the correctness of reasoning must be verified through interaction and feedback. Therefore, the proposed framework treats \emph{verification} as an integral component of both reasoning and learning.

When the AI system executes a sequence of actions, it records not only the actions themselves but also the corresponding adjustments, such as parameter changes, execution strength, timing, and environmental responses. These records allow the system to associate internal reasoning decisions with observable outcomes. Successful reasoning-action sequences reinforce the corresponding internal processes, while unsuccessful ones trigger reflection and adaptation.

Verification is also applied to internal thinking processes. A reasoning strategy is considered validated if its resulting actions consistently produce expected or desirable outcomes. Conversely, if repeated execution leads to failure or contradiction with environmental feedback, the associated reasoning process is marked as unreliable and subjected to further learning or modification. This mechanism enables the system to gradually distinguish effective reasoning patterns from ineffective ones.

For previously unseen or poorly understood problems, the system actively explores alternative reasoning and action sequences. It identifies similarities to past experiences, proposes candidate strategies, and evaluates them through interaction. Importantly, both successful and failed attempts are recorded as learning material. This exploration-driven learning allows the system to extend its internal reasoning repertoire beyond predefined scenarios.

By tightly coupling application, verification, and learning, the proposed framework ensures that internal reasoning processes are continuously grounded in real-world feedback. This design enables the AI system not only to solve known tasks more efficiently but also to incrementally develop reliable strategies for handling unknown and evolving challenges.

\section{Experimental Verification}
\label{sec_verification}

To verify the effectiveness of the proposed continuous learning framework with human-inspired internal simulation, we conduct controlled experiments on a simulated sensor-diagnosis task. The objective of the verification is to evaluate whether recording and learning internal reasoning processes can improve efficiency and interaction quality in real-time decision-making scenarios.

The experiments focus on a temperature sensor abnormality detection problem, where an AI system must identify faulty sensor behavior under uncertainty and limited observations. This task is representative of real-world monitoring applications, in which systems must reason over sequential data, request additional information when necessary, and reach conclusions through interaction rather than direct observation.

We compare multiple reasoning strategies that differ in whether and how internal reasoning processes are recorded and reused. By measuring runtime performance and the number of interactions with a large language model (LLM), we assess the impact of learning internal processes on system efficiency and adaptability.

\subsection{Experimental Setup}

The experimental environment consists of a Python-based temperature sensor simulation and an LLM-driven reasoning controller. The temperature sensor produces sequential readings over a fixed time window of eight timestamps. Under normal conditions, temperature values lie within the range of $[0, 20]^\circ$C. Abnormal sensor behavior may produce values within a wider range of $[-5, 25]^\circ$C, including cases where abnormal readings still fall within the normal range, making fault detection nontrivial.

In each simulation run, the sensor is configured to enter an abnormal state at exactly one unknown timestamp. Once the sensor becomes faulty, all subsequent readings are considered unreliable. In addition to temperature values, the simulated environment provides sensor health information, indicating whether the sensor is operating correctly. However, this information is not directly accessible unless explicitly requested by the reasoning system.

A main control program, denoted as $P_{\text{main}}$, orchestrates the interaction between the simulated environment and the LLM. The LLM used in the experiments is accessed via the DeepSeek API, with the \texttt{deepseek-chat} model serving as the reasoning core. The role of $P_{\text{main}}$ is twofold: (1) to execute actions such as querying sensor data and sensor health status, and (2) to manage internal reasoning processes, including planning before interaction and reflection after interaction.

During execution, all actions and internal reasoning steps of $P_{\text{main}}$ are recorded in a structured log. This log captures the sequence of prompts sent to the LLM, the LLM responses, the actions taken, and the observed outcomes. Depending on the reasoning strategy under evaluation, this record may be reused to guide future interactions or ignored entirely. The experiment terminates once the LLM produces a definitive conclusion identifying the timestamp at which the sensor becomes abnormal.

\subsection{Interaction and Learning Protocol}

The interaction between the reasoning system and the simulated environment follows a structured \emph{reason--act--reflect} protocol. At each stage of execution, the main controller $P_{\text{main}}$ coordinates data acquisition, reasoning, and learning-related bookkeeping, while the LLM provides high-level inference and decision guidance.

At the beginning of each run, $P_{\text{main}}$ issues an initial prompt to the LLM describing the task objective, the observable conditions, and the available actions. Based on the LLM’s response, $P_{\text{main}}$ performs the corresponding actions, such as querying temperature readings at specific timestamps or requesting sensor health information. The retrieved data are then fed back to the LLM to support subsequent reasoning steps.

The protocol distinguishes between two types of internal reasoning processes. The first occurs \emph{before} action execution and is responsible for planning, including deciding which information to request and in what order. The second occurs \emph{after} action execution and focuses on reflection, where the LLM summarizes the interaction, evaluates the effectiveness of previous steps, and adjusts its strategy if necessary. Together, these processes form a sequential reasoning loop consistent with the model described in Section~2.

Throughout the interaction, all reasoning steps, actions, and observations are recorded in a time-ordered log. This log serves as explicit learning material that captures both successful and unsuccessful reasoning patterns. Depending on the experimental configuration, the recorded log may be reused in subsequent runs to guide planning and reduce redundant interactions, or it may be ignored to simulate a non-learning baseline.

The interaction terminates when the LLM produces a definitive judgment identifying the timestamp at which the sensor becomes abnormal. If the LLM determines that the available information is insufficient, the protocol allows it to request additional data before reaching a conclusion. This flexible termination condition ensures that conclusions are reached through reasoning rather than fixed interaction limits.

\subsection{Compared Methods}

To evaluate the impact of learning internal reasoning processes, we compare three reasoning strategies that differ in whether and how prior interaction records are utilized. All methods operate within the same experimental environment and share identical task objectives and termination conditions.

\textbf{1) $M_{\text{notlearning}}$:}
This method serves as the baseline and does not leverage any learning from previous runs. All internal reasoning records, including past actions, planning steps, and reflection summaries, are discarded. Each run is treated as an independent task, and the LLM reasons solely based on the information acquired during the current interaction.

\textbf{2) $M_{\text{learning}}$:}
This method reuses recorded interaction logs from previous runs as auxiliary context. Before initiating a new interaction, the accumulated records are provided to the LLM to inform planning and reasoning. By learning from prior internal processes, the system can reduce redundant information requests and improve decision efficiency. However, the execution logic and action-selection mechanisms remain fixed.

\textbf{3) $M_{\text{noFixCodeByLearning}}$:}
This method extends $M_{\text{learning}}$ by allowing learned reasoning processes to replace predefined execution logic. In addition to reusing prior records, the LLM is encouraged to infer a local decision procedure, such as generating executable code to determine sensor abnormality without further interaction. Once such a procedure is generated and validated, the system can complete the task with minimal or no additional LLM queries.

These three methods represent increasing levels of learning integration, ranging from no learning, to learning with fixed execution, and finally to learning-driven replacement of predefined logic. Their comparison enables systematic analysis of how internal-process learning affects reasoning efficiency and interaction cost.

\subsection{Results and Analysis}

\begin{figure*}[htbp]
    \centering
    \includegraphics[width=\textwidth]{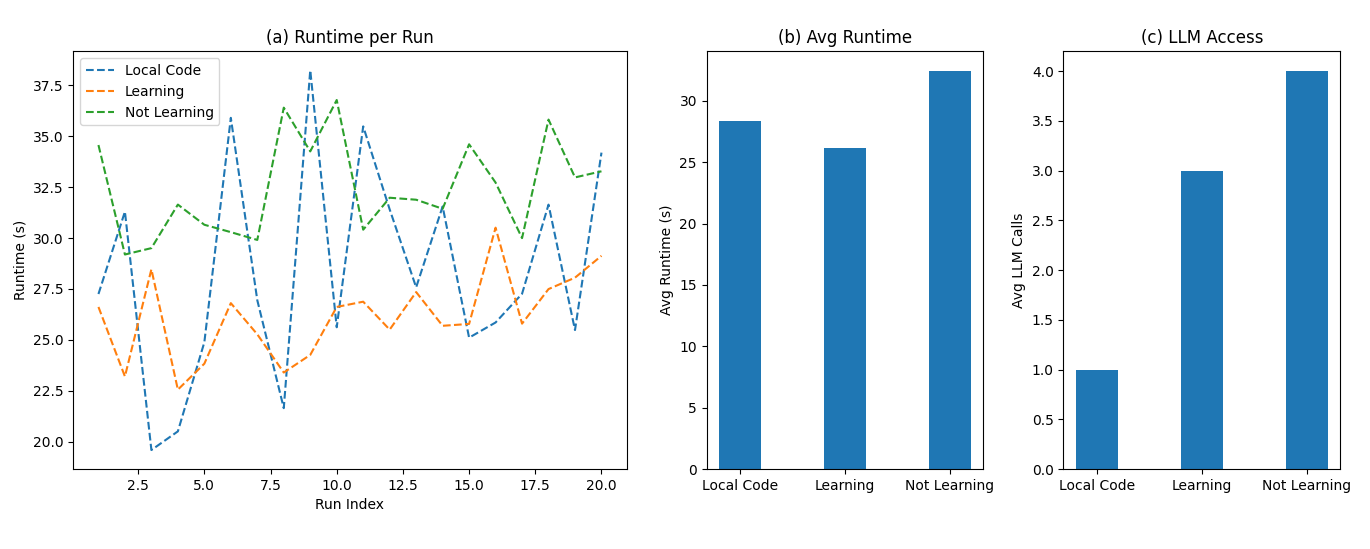}
    \caption{Performance comparison of different reasoning strategies.
    (a) Per-run runtime with dashed curves.
    (b) Average runtime across runs.
    (c) Average number of LLM interactions.
    }
    \label{fig:runtime_comparison}
\end{figure*}

Each method described in Section~3.3 is evaluated over 20 independent runs. For each run, we record the total time required to reach a definitive conclusion regarding the abnormal timestamp of the temperature sensor, as well as the number of interactions with the LLM. The aggregated results are illustrated in Fig.~\ref{fig:runtime_comparison}.

From Fig.~\ref{fig:runtime_comparison}(a) and (b), it can be observed that incorporating learning from internal reasoning processes significantly reduces overall runtime. The baseline method $M_{\text{notlearning}}$ requires an average of 32.42 seconds to reach a conclusion, whereas $M_{\text{learning}}$ achieves a lower average runtime of 26.16 seconds, reducing average runtime by 23.9\%. This improvement indicates that reusing prior reasoning and interaction records allows the system to avoid redundant information requests and reasoning steps.

The method $M_{\text{noFixCodeByLearning}}$ achieves an average runtime of 28.37 seconds, which lies between the other two methods. Although this approach benefits from learning-driven optimization, the additional time required for the LLM to generate and validate executable decision logic partially offsets the reduction in interaction cost. This result highlights a trade-off between upfront reasoning complexity and long-term execution efficiency.

Fig.~\ref{fig:runtime_comparison}(c) further compares the average number of LLM interactions required by each method. The baseline $M_{\text{notlearning}}$ requires approximately four LLM interactions per run, reflecting repeated data requests and incremental reasoning. By contrast, $M_{\text{learning}}$ reduces the average number of interactions to three by leveraging prior records to request multiple data sources jointly. Notably, $M_{\text{noFixCodeByLearning}}$ requires only a single interaction on average, as the learned local decision procedure enables the system to complete the task without further LLM queries.

These results demonstrate that recording and learning internal reasoning processes can substantially improve interaction efficiency. Moreover, enabling learned processes to replace fixed execution logic further reduces dependency on external reasoning services. Together, these findings validate the proposed framework’s ability to optimize both reasoning behavior and system-level performance through continuous learning of internal processes.

\subsubsection{Statistical Significance Analysis}

To further evaluate whether the observed performance differences are statistically significant, we conduct pairwise statistical tests on the total runtime across different methods. Since each method is evaluated over independent runs and equal variance cannot be guaranteed, we employ a two-sided Welch’s $t$-test, which does not assume homogeneity of variance.

Let $\mu_i$ and $\mu_j$ denote the mean runtimes of methods $i$ and $j$, respectively. For each pair of methods, the null hypothesis $H_0$ assumes no difference in mean runtime ($\mu_i = \mu_j$), while the alternative hypothesis $H_1$ assumes a significant difference ($\mu_i \neq \mu_j$). The significance level is set to $\alpha = 0.05$.

The comparison between $M_{\text{learning}}$ and $M_{\text{notlearning}}$ yields a statistically significant reduction in runtime, with $p < 0.05$, indicating that incorporating learning from internal reasoning processes leads to a measurable efficiency improvement. This result confirms that reusing prior reasoning and interaction records effectively reduces redundant computation and interaction overhead.

When comparing $M_{\text{noFixCodeByLearning}}$ with $M_{\text{notlearning}}$, the test also indicates a significant difference ($p < 0.05$), demonstrating that replacing fixed execution logic with learned procedures substantially improves system efficiency. In contrast, the difference between $M_{\text{learning}}$ and $M_{\text{noFixCodeByLearning}}$ is not statistically significant at the same confidence level, suggesting that the additional overhead introduced by code generation partially offsets the gains from reduced LLM interaction.

Overall, the statistical analysis supports the conclusion that learning internal reasoning processes provides significant performance benefits, while more aggressive replacement of fixed logic introduces a trade-off between reasoning cost and interaction reduction. These findings reinforce the effectiveness and robustness of the proposed continuous learning framework.

\section{Conclusion}
\label{sec_conclusion}

This paper presented a human-inspired continuous learning framework that enables adaptive AI systems to learn not only task-specific content but also internal reasoning activities, scheduling mechanisms, and learning strategies. By integrating sequential reasoning with parallel learning, the proposed approach embeds learning directly within the reasoning--action loop, allowing internal cognitive structures to evolve during execution. In addition, the framework supports controlled replacement of predefined logic with learned procedures and introduces a hierarchical learning-to-learn mechanism for meta-level adaptation.

Experimental results on a temperature sensor abnormality detection task demonstrated that incorporating internal-process learning improves reasoning efficiency, achieving a 23.9\% reduction in average runtime while reducing interaction overhead.

Future work will explore the incorporation of additional human-inspired cognitive mechanisms into AI systems to further enhance adaptability and structural learning capabilities. In particular, extending the framework to more complex embodied agents and intelligent robotic systems may enable more human-like reasoning and decision-making in dynamic environments.


\ifCLASSOPTIONcaptionsoff
  \newpage
\fi

\bibliographystyle{IEEEtran}
\bibliography{ref}

%

\begin{IEEEbiography}{Hong Su}
  received the MS and PhD degrees, in 2006 and 2022, respectively, from Sichuan University, Chengdu, China. He is currently a researcher of Chengdu University of information Technology Chengdu, China. His research interests include blockchain, cross-chain and smart contract.
\end{IEEEbiography}




\end{document}